\documentclass{article}

\usepackage[preprint]{corl_2021} 

\usepackage{microtype}
\usepackage{graphicx}
\usepackage{subfigure}
\usepackage{booktabs} 

\usepackage{hyperref}

\usepackage{enumitem}

\usepackage[font=small,labelfont=bf]{caption}
\usepackage{amssymb}
\usepackage{algorithm,algorithmic}
\usepackage{amsmath,bm}
\usepackage{amsthm}
\usepackage{mathrsfs} 
\usepackage{float}

\newcommand{\tabincell}[2]{\begin{tabular}{@{}#1@{}}#2\end{tabular}}

\usepackage[english]{babel}
\usepackage{blindtext}
\usepackage{multicol}

\title{Interpretable Model-based Hierarchical Reinforcement Learning using Inductive Logic Programming}

\author{
  Duo Xu\\
  Department of Electrical and Computer Engineering\\
  Georgia Institute of Technology\\
  \texttt{dxu301@gatech.edu} \\
  \AND
  Faramarz Fekri\\
  Department of Electrical and Computer Engineering\\
  Georgia Institute of Technology\\
  \texttt{faramarz.fekri@ece.gatech.edu} \\
}

\begin{document}
\maketitle


\begin{abstract}
Recently deep reinforcement learning has achieved tremendous success in wide ranges of applications. However, it notoriously lacks data-efficiency and interpretability. Data-efficiency is important as interacting with the environment is expensive. Further, interpretability can increase the transparency of the black-box-style deep RL models and hence gain trust from the users. In this work, we propose a new hierarchical framework via symbolic RL, leveraging a symbolic transition model to improve the data-efficiency and introduce the interpretability for learned policy. This framework consists of a high-level agent, a subtask solver and a symbolic transition model. Without assuming any prior knowledge on the state transition, we adopt inductive logic programming (ILP) to learn the rules of symbolic state transitions, introducing interpretability and making the learned behavior understandable to users. In empirical experiments, we confirmed that the proposed framework offers approximately between 30\% to 40\% more data efficiency over previous methods.
\end{abstract}



\vspace*{-5pt}
\section{Introduction}
\vspace*{-5pt}
Reinforcement learning (RL) methods have established themselves as the state of the art for solving complex sequential decision making problems in games and robotics, leading to significant impacts in practical applications \citep{van2015learning,mnih2015human,kumar2016learning,kumar2016optimal,akkaya2019solving,falco2018policy}. 
However, since the environment model and reward function are unknown initially, RL methods mostly rely on random exploration to collect rewards and then improve their current policy accordingly. Therefore, RL methods are notoriously sample inefficient, requiring a large amount of interactions with the environment before learning policies better than random exploration. This problem becomes more severe in long-horizon tasks \citep{andrychowicz2020learning}. Another problem in deep RL is the lack of interpretability \citep{lyu2019sdrl,puiutta2020explainable}. The learned behavior based on the black-box neural network is nontransparent and difficult to explain and understand. The goal of interpretability is to describe the internals of a system or learned behavior in a way that they are readable and verifiable by humans. In some  real-world applications of RL, it is instrumental to make the system behavior interpretable to human, so as to make the system more reliable and user-friendly \citep{israelsen2019dave,dovsilovic2018explainable}.

In this work we propose to use model-based Hierarchical RL (HRL) via inductive logic programming (ILP) to tackle problems mentioned above. First, HRL is a promising approach to reducing sample complexity and scaling RL to long-horizon tasks \citep{peng2017deeploco}. The idea is to use a high-level policy to generate a sequence of high-level goals, forming subtasks, and then use low-level policies to generate sequences of actions to solve every subtask. By abstracting many details of the low-level states, the high-level policy can efficiently plan over much longer time horizons, reducing sample complexity in many tasks. In addition, because of explicit knowledge representation in the hierarchical formulation, performing reasoning and planning on high-level goals become an effective way to introduce interpretability into deep RL. Different from previous work on HRL with symbolic planning \citep{leonetti2016synthesis,lyu2019sdrl,illanes2020symbolic}, we do not need any prior knowledge on symbolic transitions in the high-level part. Instead, by leveraging the inductive logic programming (ILP) \citep{evans2018learning,garcez2019neural,payani2019learning}, we adopt the model-based RL \citep{janner2019trust,kurutach2018model}, which learns a transition model of the high-level symbolic states via predicate logic language in ILP and utilize this model to generate subtask sequences improving data-efficiency and interpretability. 
In contrast to previous works on ILP, we propose to incorporate refinement operations to generate clauses more efficiently, and hence better utilizing real transition experiences.


As a result, the proposed framework provides the following benefits: (i) it improves the sample efficiency by leveraging the hierarchical learning framework and the symbolic state transition model, (ii) it introduces interpretability into deep RL via the learned symbolic state transition rules, and (iii) it provides the compositional generalization via the ILP. The effectiveness of the proposed method is verified by empirical experiments, compared with previous methods such as HRL \citep{kulkarni2016hierarchical}.

\vspace*{-5pt}
\section{Preliminary}
\vspace*{-5pt}
In this section, we establish relevant notation and review key aspects of symbolic reinforcement learning.

\vspace*{-5pt}
\subsection{Reinforcement Learning}
\vspace*{-5pt}
For the purposes of this work, we will say that the environment with which an RL agent interacts is formalized as a Markov Decision Process (MDP) $M=(S, A, r, p, \gamma)$, where $S$ is the state space, $A$ is the set of actions, $r:S\times A\to\mathbb{R}$ is the corresponding reward function, $p(s_{t+1}|s_t, a_t)$ is the state transition probability given any state-action pair, and $\gamma\in[0, 1)$ is the discount factor. A policy for $M$ is defined as a probability distribution $\pi(a|s)$ representing the probability of the agent taking action $a$ given that its current state is $s$. Therefore, the RL problem is to find the optimal policy $\pi^*$ maximizing the expected discounted future reward obtained from all states $s\in S$ \citep{sutton1999between}:
\begin{equation}
    \pi^*=\arg\max_{\pi}\sum_{s\in S}v_{\pi}(s) \nonumber
\end{equation}
where $v_{\pi}(s)$ is defined as the value function, approximating the expected discounted future reward obtained when starting at state $s\in S$ following the policy $\pi$, i.e., 
\begin{equation}
    v_{\pi}(s)=\mathbb{E}_{\pi}\bigg[\sum_{t=0}^{\infty}\gamma^tr_t\bigg|s_0=s\bigg] \nonumber 
\end{equation}

In this work, our method is built on Q learning \citep{watkins1992q}, which is an RL approach that learns optimal policies (in the limit) by using sampled experiences to estimate the optimal q-function $q^*(s,a)$ for every state $s\in S$ and action $a\in A$. The optimal q-function $q^*(s,a)$ is equal to the expected discounted future reward received by performing action $a$ in state $s$ and following an optimal policy. Given an experience tuple $(s,a,r',s')$, the q-value estimate $\tilde{q}(s,a)$ is updated as follows
\begin{equation}
    \tilde{q}(s,a)\longleftarrow\bigg(r'+\gamma\max_{a'\in A}\tilde{q}(s,a')\bigg) \nonumber
\end{equation}
Here the optimal policy $\pi^*$ can be easily derived from $q^*(s,a)$ by selecting the action $a\in A$ with the largest q-value under the current state $s\in S$. In order to explore the environment, the $\epsilon$-greedy exploration strategy is often used in Q-learning, selecting the random action with probability $\epsilon$ and the action with the largest $\tilde{q}(s,\cdot)$ value with probability $1-\epsilon$.

\vspace*{-5pt}
\subsection{The Option Framework}
\vspace*{-5pt}
The options framework is a framework for defining and solving semi-Markov Decision Processes (SMDPs) with a type of macro-action called an option \citep{sutton1999between}. The including options into an MDP problem turns it into an SMDP problem, because actions are dependent not only on the previous state but also on the identity of the currently active option, which could have been initiated many time steps before the current time. 

An {\em option} $o$ is temporally extended course of action consisting of three components: a policy $\pi_o:S\times A\to[0,1)$, a termination condition $\beta_o:S\to[0, 1]$, and an initial set $I_o\in S$. An option $(I_o, \pi_o, \beta_o)$ is available in state $s_t$ if and if only $s_t\in I_o$. After the option is taken, a sequence of actions is selected according to $\pi_o$ until the option $o$ is terminated with the probability of the termination condition $\beta_o$. With the introduction of options, we can formulate the decision-making as a hierarchical process with two levels, where the high level is the option level (also termed as task level) and the lower level is the action (sub-task) level. Markovian property exists among different options at the option level. A crucial benefit of using options is that they can be composed in arbitrary ways.

\vspace*{-5pt}
\subsection{Inductive Logic Programming}
\vspace*{-5pt}
Logic programming languages are a class of programming languages using logic rules rather than imperative commands. By adopting the programming language of {\em DataLog} \citep{koller2007introduction}, we define our logic language as below. Having predicate names (predicates), constants, and variables as three primitives, the predicate name is defined as a relation name, and a constant is termed as an entity. An {\em atom} $\alpha$ is defined as a predicate followed by a tuple $p(t_1,\ldots,t_n)$, where $p$ is an $n$-ary predicate and $t_1,\ldots,t_n$ are terms, i.e., variables or constants. For example, the atom $on(X, \text{ground})$, denotes the predicate called $\text{on}$ with $X$ as variable and $\text{ground}$ as constant. If all terms in an atom are constants, this atom is called a {\em ground atom}. In this work the set of all ground atoms is denoted as $G$. A predicate, which can be defined by a set of ground atoms, is called an {\em extensional predicate}. Further, a {\em clause} is defined as a rule in the form of $\alpha\leftarrow\alpha_1,\ldots,\alpha_n$, where $\alpha$ is the {\em head atom}, and $\alpha_1,\ldots,\alpha_n$ are {\em body atoms}. The predicates defined by clauses are termed as {\em intensional predicates}.

Inductive logic programming (ILP) is a task to derive a definition (set of clauses) of some intensional predicates, given some positive examples and negative examples \citep{koller2007introduction,evans2018learning}. Conducting ILP with differentiable architectures has been investigated in many previous work \citep{evans2018learning,rocktaschel2017end,dong2019neural,payani2019learning}. In this work, we adopt $\partial$ILP \citep{evans2018learning} as the base method. 
With the differentiable deduction, the system can be trained with gradient-based methods. The loss value is defined as the cross-entropy between the confidence of predicted atoms and the ground truth. Compared with traditional ILP methods, $\partial$ILP has advantages in terms of robustness against noise and ability to deal with fuzzy data \citep{evans2018learning}.

\vspace*{-5pt}
\section{Related Work}
\vspace*{-5pt}
{\bf Interpretability} There have been a lot of recent papers investigating the interpretability in deep learning \citep{doshi2017towards,gilpin2018explaining,roscher2020explainable}. Making the black-box deep neural network explainable to human is also an active research area, having strong practical impact \citep{tjoa2020survey}. There are some papers studying interpretable RL from the perspective of programming synthesis \citep{bunel2018leveraging,verma2018programmatically}. However, many unsolved problems on interpretable RL are left to be investigated.

{\bf Symbolic RL} Some recent papers study the interpretability of RL by integrating symbolic planning \citep{leonetti2016synthesis,lu2018robot,yang2018peorl,lyu2019sdrl,illanes2020symbolic}, which inherit the interpretability of symbolic planning with symbolic knowledge. However, all of them require prior knowledge on action description, i.e., the effects of symbolic actions on the symbolic state representations, and they only conduct model-free RL in the symbolic state space. In this work, this prior knowledge is not required, and the learned symbolic transition model can enable the model-based RL in the high level and improve both the data-efficiency and interpretability.

\vspace*{-5pt}
\section{Methodology}
\vspace*{-5pt}
We first give an overview of the proposed framework. In addition to the primitive state and action spaces $(S, A)$, following previous works \citep{leonetti2016synthesis,lyu2019sdrl,illanes2020symbolic}, we assume that the agent has access to several objects $\mathcal{O}$ and a set of predicates $\mathcal{P}$ (i.e. relationships over objects) implemented by a human expert, which help us to formulate a hierarchical SMDP to solve the problem.
In contrast to the previous work, we use inductive logic programming (ILP) to the learn symbolic transition model in the high level portion, improving sample efficiency and task-level interpretability.


\begin{figure}
    \centering
    \includegraphics[width=2.2in]{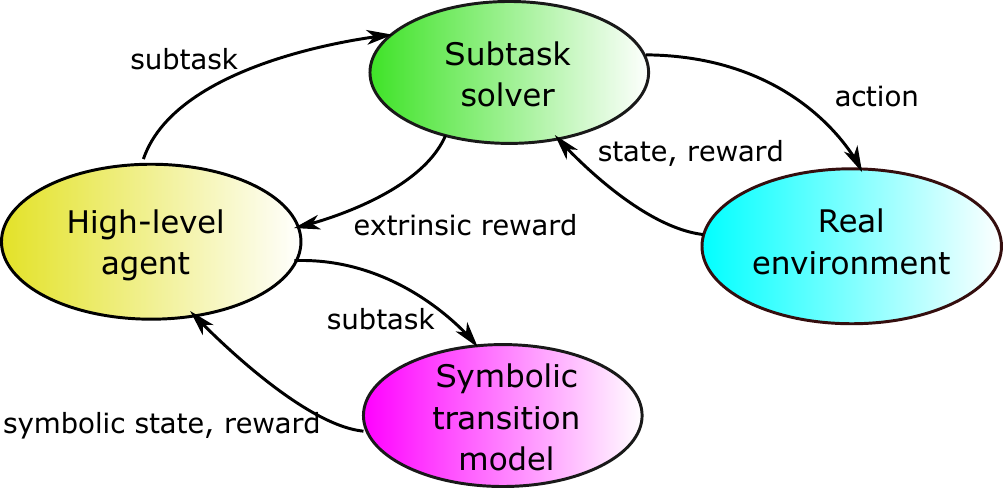}
    \caption{Diagram of Hierarchical Framework}
    \label{fig:hrl}
\end{figure}

\vspace*{-5pt}
\subsection{Predicates in ILP}
\vspace*{-5pt}
In this section, we are going to formally define the predicates, formulating symbolic states in the hierarchical SMDP. The high-level learning is conducted over symbolic states formulated by the objects and predicates. The set of objects, denoted as $\mathcal{O}$, formulates abstractions of environmental states. Predicates, denoted as $\mathcal{P}$, are Boolean-valued truth statements corresponding to goals, relationships of objects, events and properties of the environment. The predicates in $\mathcal{P}$ are classified into subgoals $\mathcal{P}_G$, environmental properties $\mathcal{P}_C$ and events $\mathcal{P}_E$. Specifically, every predicate in $\mathcal{P}_G$ indicates whether certain subgoal is achieved or not. For example, $\mathcal{P}_G$ can consist of AchieveObj$(o), \forall o\in\mathcal{O}$, indicating that the high-level agent reaches certain object $o$ by satisfying some preconditions. Predicates in $\mathcal{P}_C$ represent properties of the environment which are constant during the learning, e.g., RoomHasKeyColor(X,C), which is a predicate indicating that 'room X contains a key in color C'. Finally a predicate in $\mathcal{P}_E$ denotes the occurrence of the event during the interaction of RL agent with the environment, e.g., hasKeyColor(C), representing the event that the agent has obtained the key in color C. In other words, this predicate will be true only when the agent has the key with color C. The values of predicates in $\mathcal{P}$ are given by the environment as auxiliary information for the RL agent.

\vspace*{-5pt}
\subsection{Hierarchical SMDP}
\vspace*{-5pt}
\label{sec:framework}
We assume that every environmental state $s\in\mathcal{S}$ has a corresponding symbolic state $\hat{s}$ which is defined as the subset of predicates in $\mathcal{P}$ holding true in that state $s$. The set of all the possible symbolic states is denoted as $\mathcal{\hat{S}}$. We also assume the existence of a labeling function $L(s)$ which produces the symbolic state $\hat{s}$ corresponding to the environmental state $s$. There is no need to learn this labeling function $L(s)$, since the valuation of predicates is directly given by the environment. 

The proposed framework is built on a hierarchical semi-Markov Decision Process (SMDP), learning the options and planning over them, as shown in Figure \ref{fig:hrl}. The high-level part of the SMDP is defined over the symbolic state $\mathcal{\hat{S}}$ with subgoals in $\mathcal{P}_G$ as actions, which forms a symbolic MDP $\hat{\mathcal{M}}:=(\hat{\mathcal{S}}, \mathcal{P}_G, R_H, T_H, \gamma)$, where $R_H$ and $T_H$ are the reward function and the symbolic state transition function, respectively. The low-level part is to learn options over the environmental MDP $\mathcal{M}=(\mathcal{S}, \mathcal{A}, R, T, \gamma)$, solving subtasks assigned by the high-level agent. Therefore, the hierarchical SMDP can be defined as the product of the environmental and symbolic MDPs, i.e., $\mathcal{M}_{\text{SMDP}}=(\mathcal{S}\times\hat{\mathcal{S}}, \mathcal{A}\times\mathcal{P}_G. R_{\text{SMDP}}, T\times T_H, \gamma)$. In the high level, the transition tuple is defined as $(\hat{s},p,\hat{s}')$ where $\hat{s}$ and $\hat{s}'$ are current and next symbolic states, and $p$ is the subgoal selected by the agent. Specifically, the transition where the subgoal $p$ is achieved in $\hat{s}'$ is denoted as {\it successful} transition, while the transition where subgoal $p$ is not achieved is denoted as {\it unsuccessful} transition, where $p$ grounded in $\hat{s}'$ is False.


Specifically, based on the predicate information, the high level portion of the SMDP, modeled by the symbolic MDP $\hat{\mathcal{M}}$, is solved by the model-based reinforcement learning (MBRL) method. Different from previous HRL methods, by leveraging ILP, we propose to learn a symbolic state transition model described by logic rules. As shown in Figure \ref{fig:hrl}, the high-level agent solves the symbolic MDP $\hat{\mathcal{M}}$ by interacting with both the symbolic transition model and the real environment, which can reduce sampling complexity significantly as other MBRL methods \citep{moerland2020model}. 

As previous papers on symbolic planning \citep{gopalan2017planning,lyu2019sdrl,illanes2020symbolic}, for every subgoal $p\in\mathcal{P}_G$, the state transitions $T_H$ can be partitioned into {\it pre-conditions} and {\it effects}, denoted as pre($p$) and eff($p$) respectively. The pre-conditions represent the prerequisites for achieving the subgoal, and the effects denote the environmental events triggered after achieving the subgoal. Different from previous works, here we propose to use ILP to learn logic rules to describe preconditions and effects of achieving subgoals.

In the low level portion of $\mathcal{M}_{\text{SMDP}}$, the subtask solver in Figure \ref{fig:hrl} learns options in the environmental MDP $\mathcal{M}$. Every option is associated with a subgoal in $\mathcal{P}_G$, and the target of option is to achieve that subgoal. The subtask is a variable-length sequence of actions which achieve a subgoal following the corresponding option. Every subtask is defined by a tuple of subgoals $(p, p')$, where $p$ denotes the initial subgoal satisfied by the initial state and $p'$ is the subgoal to be achieved next.

\vspace*{-5pt}
\subsection{Options and Rewards}
\vspace*{-5pt}
For every subgoal $p\in\mathcal{P}_G$, there is an associated option $o_p=(\mathcal{I}_{o_p}, \pi_{o_p}, \beta_{o_p})$, which is to achieve the subgoal $p$ in the low level of the hierarchical SMDP. Specifically, the initial set of every option is set to the state space of the environmental MDP directly, i.e., $\mathcal{I}_{o_p}=\mathcal{S}$. Each $o_p$ has its own policy $\pi_{o_p}$ whose goal is to reach the states satisfying the subgoal $p$ denoted as $\mathcal{S}_p$, i.e., $\mathcal{S}_p:=\{s\in\mathcal{S}| p\in L(s)\}$. Options are learned in the environmental MDP $\mathcal{M}$, which terminates only when any state in $\mathcal{S}_p$ is reached, defining the function $\beta_{o_p}$. We have that $\beta_{o_p}(s)=1$ only when $s\in\mathcal{S}_p$, otherwise $\beta_{o_p}(s)=0$. The learning method for policies of options is the conventional RL method, such as PPO \citep{schulman2017proximal} or Q-learning \citep{watkins1992q}. 

In the proposed framework, as shown in Figure \ref{fig:hrl}, the reward functions for the low and high levels of the hierarchical SMDP are formulated as {\em intrinsic reward} $r_i$ and {\em extrinsic reward} $r_e$, respectively \citep{yamamoto2018hierarchical,le2018hierarchical,illanes2020symbolic}. Hence in the definition of the SMDP, we have $R_{\text{SMDP}}:=r_i\times r_e$. The options in the low level are trained by intrinsic rewards which have pseudo-rewards to encourage the agent to reach the specified subgoal. Given every subgoal $p\in\mathcal{P}_G$, the {\em intrinsic reward} function in terms of transition $(s, a, s')$ is defined as 
\begin{equation}
    r_i(s, a, s'; p)=\begin{cases}
    \eta\hspace{49pt}p\in L(s') \\
    r(s, a, s')\hspace{15pt}\text{otherwise}
    \end{cases}\label{intrinsic}
\end{equation}
where $\eta$ is a positive number to encourage the agent to achieve the subgoal $p$, and $r$ is the reward for the valid movement or environmental reward. Besides, the high-level agent conducts Q learning with {\em extrinsic reward} which is defined in terms of a subtask $(p, p'), \forall p, p'\in\mathcal{P}_G$, as below,
\begin{equation}
    r_e(p, p')=\begin{cases}
    R(p, p')\text{,}\hspace{20pt}0.9<t(p, p') \\
    -\xi_0\text{,}\hspace{35pt}0<t(p, p')<0.9 \\
    -\xi_1\text{,}\hspace{35pt}t(p, p')<0.9\text{ and }N<n(p, p')\\
    \end{cases}\label{extrinsic}
\end{equation}
where $t(p, p')$ denotes the success rate of $(p, p')$, and $n(p, p')$ is the number of times that the subtask has been tried thus far. Specifically, $0<\xi_0\ll\xi_1$ refers to the penalty for immature and unlearnable subtasks, respectively. Immature subtasks refer to those without sufficient training, and unlearnable ones are those too difficult to solve. We penalize unlearnable subtasks more heavily than immature ones. If the subtask can be solved robustly, the extrinsic reward is set to be $R(p, p')$ which is the accumulated environmental rewards $r$ when the agent is achieving the subgoal $p'$ starting from any initial state $s_0$ satisfying $p\in L(s_0)$.

\vspace*{-5pt}
\subsection{Learning Symbolic Transition Model via ILP}
\vspace*{-5pt}
\label{sec:ilp}
Different from previous work on HRL \citep{kulkarni2016hierarchical,pateria2021hierarchical} and symbolic planning \citep{lyu2019sdrl,illanes2020symbolic}, in this work we use ILP to learn a symbolic state transition model consisting of learned clauses. 
The ILP method adopted is a refinement-based $\partial$ILP which integrates refinement operation \citep{cropper2020turning} and $\partial$ILP \citep{evans2018learning} together. This ILP method can lift many limitations in $\partial$ILP, such as the maximum number of predicates in the body of clauses, and the requirements for rule templates. The general idea is that by leveraging the generality of clauses and real transition experiences, we use {\it refinement operations} \citep{cropper2020turning,cropper2021learning} to form clauses, and hence avoiding many meaningless clauses.
Specifically, the refinement operator takes a clause and returns weakened (more specified) clauses, which is introduced with details in Section \ref{sec:app_refinement} in Appendix. 

Assume that the union set of input and target predicates is denoted as $\mathcal{F}$. The original $\partial$ILP operates on the valuation vectors whose space is $E=[0, 1]^{|\mathcal{F}|}$, each element of which represents the confidence of a related atom (grounded predicate) in $\mathcal{F}$. Denote $\bm{e}_0$ as the valuation (true or false) of all the atoms in $\mathcal{F}$ grounded by the input state $s\in\mathcal{S}$. Learning $\partial$ILP model is to search clauses from the set of generated clauses denoted as $\mathcal{C}$, which can entail positive examples and preclude negative examples \citep{evans2018learning}.  Different from original $\partial$ILP, we introduce refinement operation to generate clauses. 

We define a mapping $d_{\bm{\phi}}:E\to E$ with parameters $\bm{\phi}$ which denotes deduction of facts $\bm{e}_0$ using weights $\bm{\omega}$ associated with all the generated clauses $\mathcal{C}$. The mapping $d_{\bm{\phi}}$ consists of repeated applications of single-step deduction function $g_{\bm{\phi}}$ which is described as below,
\begin{equation}
    d_{\phi}^t(\bm{e}_0)=\begin{cases}
    g_{\phi}(d_{\phi}^{t-1}(\bm{e}_0))\hspace{15pt}\text{if }t>0 \\
    \bm{e}_0\hspace{60pt}\text{if }t=0
    \end{cases}
    \label{ilp}
\end{equation}
where $t$ is the deduction step, and $g_{\bm{\phi}}$ represents one-step deduction of clauses in $\mathcal{C}$ weighted by trainable weights $\bm{\omega}$. Defining probabilistic sum $\oplus$ as $\bm{a}\oplus\bm{b}=\bm{a}+\bm{b}-\bm{a}\odot\bm{b}, \forall \bm{a},\bm{b}\in E$, we can express the operation of single-step deduction as below
\begin{equation}
    g_{\bm{\phi}}(\bm{e})=\bigg(\sum_i^{\oplus}\sum_j\omega_{i,j}f_{i,j}(\bm{e})\bigg)+\bm{e}_0 \label{ilp2}
\end{equation}
where function $f_{i,j}$ implements one-step deduction using $j$th definition of $i$th clause in $\mathcal{C}$, with $\bm{\omega}_{i,\cdot}$ as its trainable weight \citep{evans2018learning,jiang2019neural}. For the specific $i$th clause, we can constrain the sum of its weights to be $1$ by letting $\bm{\omega}_i=\text{softmax}(\bm{\phi}_i)$, where $\bm{\phi}_i$ are related parameters to be trained. Then the transition model can be denoted as $\tilde{T}_{H,\bm{\phi}}$, parameterized by the same parameters as ILP model. The training of $\tilde{T}_{H,\bm{\phi}}$ is conducted by regression on real transition tuples $(\hat{s},p,\hat{s}')$. Note that in unsuccessful transitions $\hat{s}$ is same as $\hat{s}'$. The input predicates are grounded by $\hat{s}$ and the target predicates are grounded by $\hat{s}'$. The loss function is the binary cross entropy between predicted and real target atoms.

In the proposed framework, we use this proposed ILP method to learn logic rules to describe pre-conditions and effects separately. Specifically, given each subgoal $p\in\mathcal{P}_G$, for learning preconditions (pre($p$)), the target predicate (to be learned) is the subgoal $p$ and the input predicates (used as background information in ILP learning) are properties and events of the environment, i.e., $\mathcal{P}_E\cup\mathcal{P}_C$. However, for learning effects (eff($p$)), the target predicates are event predicates in $\mathcal{P}_E$ whereas the input predicates are subgoal and environmental properties, i.e., $\{p\}\cup\mathcal{P}_C$. 


\vspace*{-5pt}
\subsection{Algorithm}
\vspace*{-5pt}
In every training episode, based on the setting shown in Figure \ref{fig:hrl}, the high-level agent selects the subgoal and forms a subtask which is then assigned to the sub-task solver, until the task is completed or timeout step is achieved. We also use ILP to learn a symbolic state transition model which uses logic rules to describe state transitions $T_H$ in symbolic MDP $\hat{\mathcal{M}}$. Then, the high-level agent can interact with the real environment (via sub-task solver) and the simulated transition model in an alternating way, resulting in significant reduction in the sample complexity. In the low level the subtask solver solves every subtask by learning the corresponding option in environmental MDP $\mathcal{M}$. 

The details of the proposed method are illustrated in Algorithm \ref{alg} in Appendix. 


\vspace*{-5pt}
\section{Experiments}
\vspace*{-5pt}
We evaluate the proposed approach in two environments. The first environment is the modified room environment \citep{le2018hierarchical,abel2020value} which has discrete state and action spaces. The second environment is a variant of OpenAI's Safety Gym \citep{ray2019benchmarking} which has continuous state and action spaces, called Robot Navigation. 

In both environments, we quantitatively verify the advantages of the proposed method only in sample efficiency and generalization. The high-level interpretability is automatically given by the learned logic rules in the symbolic transition model, due to the human readability of logic rules.  We never consider any RL methods without utilizing hierarchy or symbolic information of the environment as baselines, since the performances of these methods are obviously worse and not meaningful for comparisons. 


\vspace*{-5pt}
\subsection{Room Environment}
\begin{figure}[ht]
\centering
\subfigure[Training Map]{
    \includegraphics[width=1.1in]{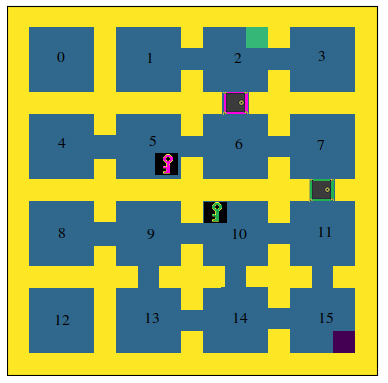}
    \label{fig:room1_1}
}
\subfigure[Learning Curves]{
    \includegraphics[width=1.39in]{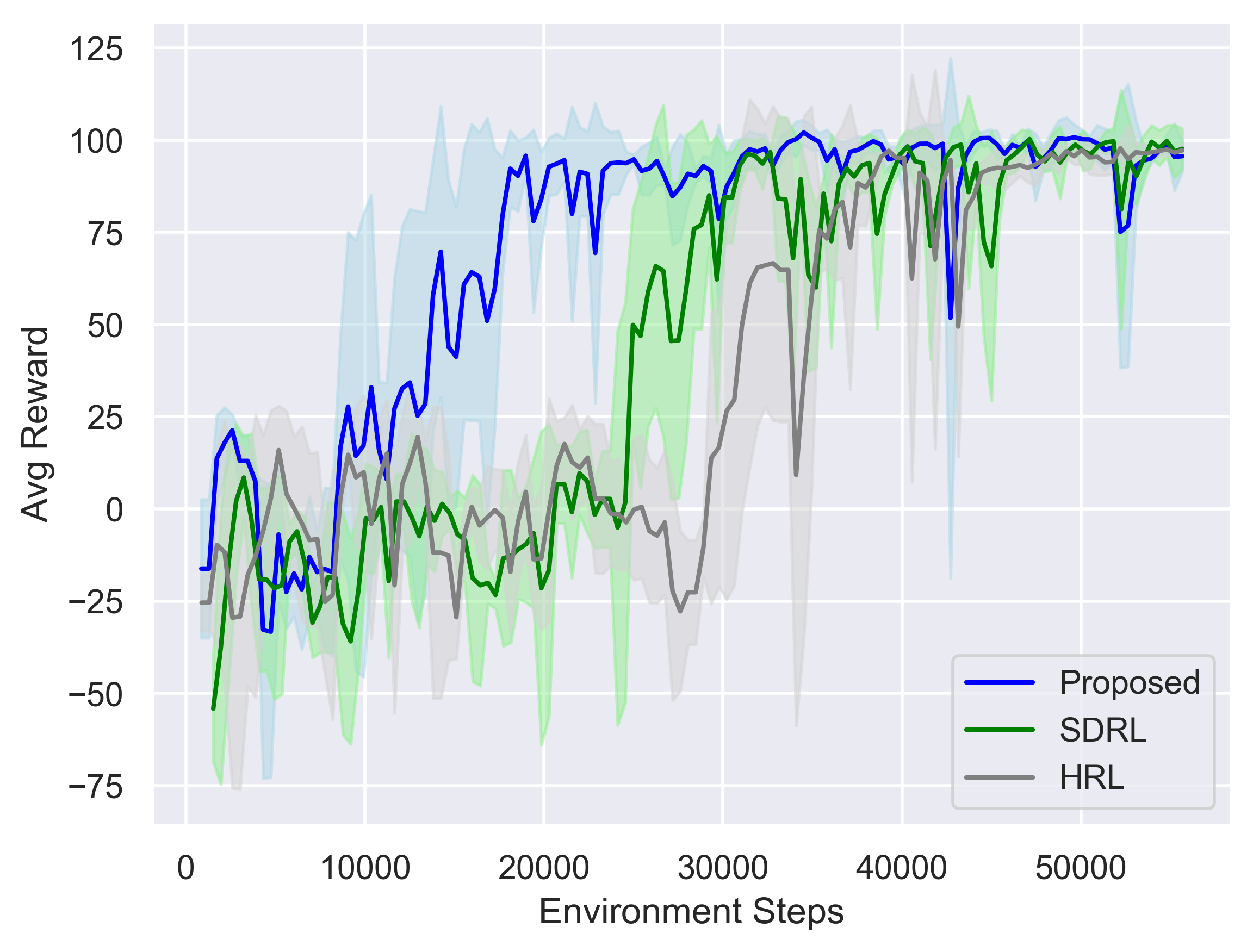}
    \label{fig:room1_2}
}
\subfigure[Subtask Success Rate]{
    \includegraphics[width=1.39in]{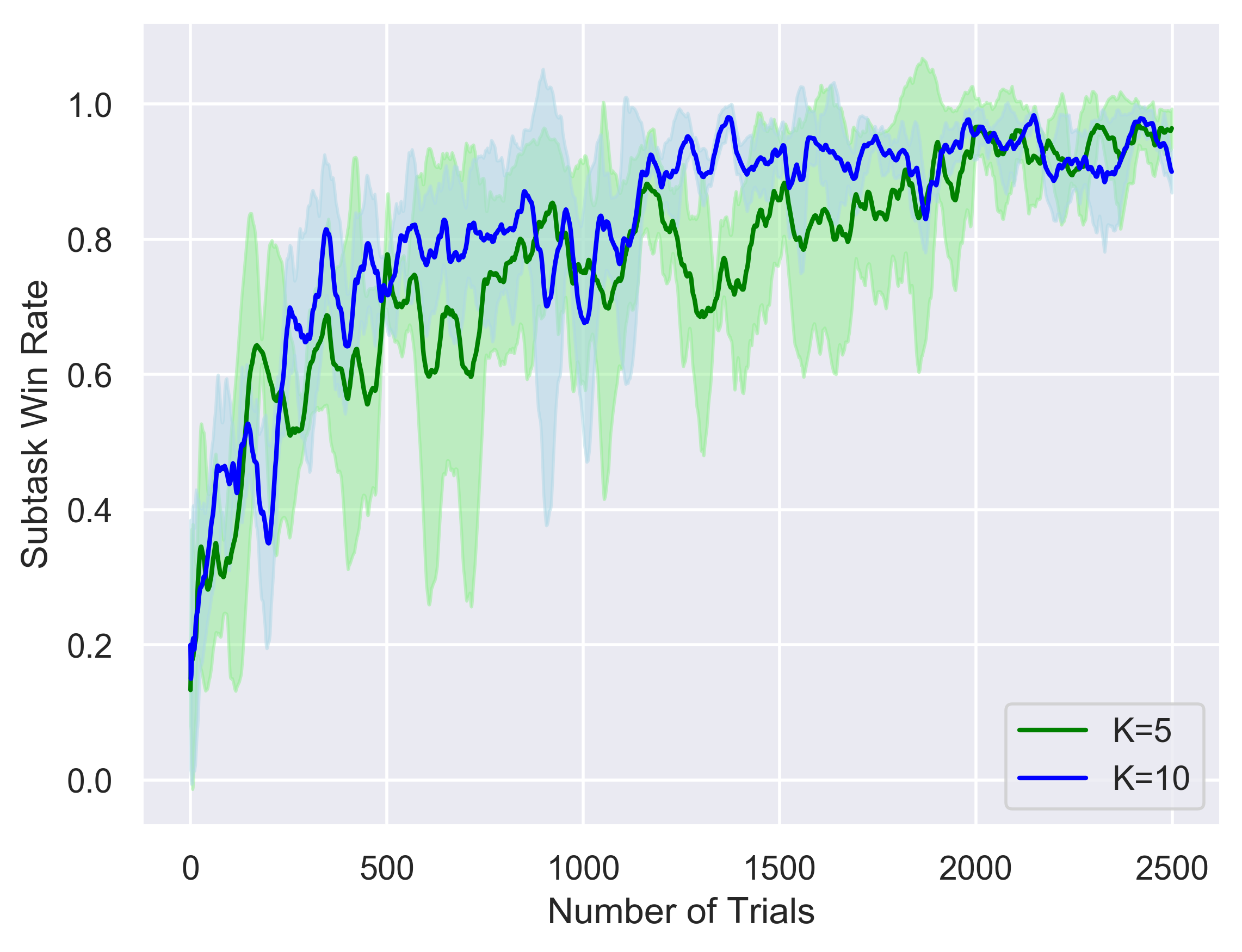}
    \label{fig:room1_3}
}
\vspace*{-5pt}
\caption{Room Environment for Training and Learning Performance. There are two pairs of locks and keys in red and green. The robot starts at black grid and targets at the green grid}
\label{fig:room1}
\vspace*{-5pt}
\end{figure}

\begin{figure*}[ht]
\centering
\subfigure[Test 1 Map]{
    \includegraphics[width=1.15in]{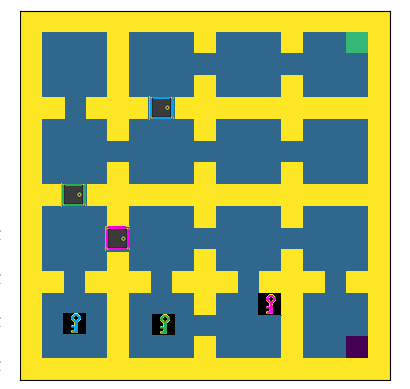}
    \label{fig:room2_2}
}
\subfigure[Test 2 Map]{
    \includegraphics[width=1.1in]{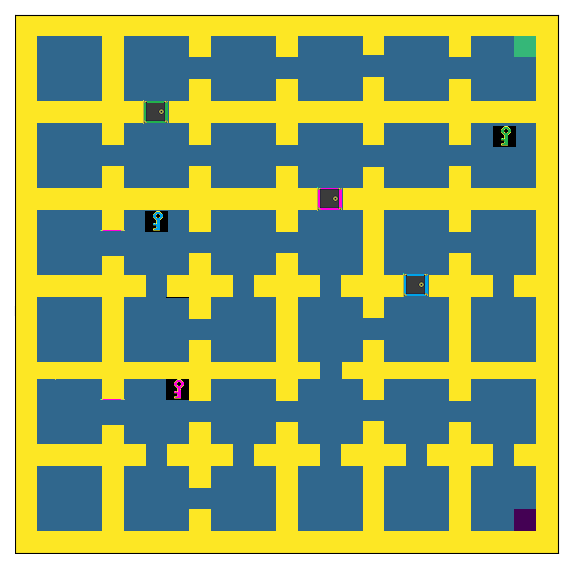}
    \label{fig:room2_3}
}
\subfigure[Test 1 Curve]{
    \includegraphics[width=1.39in]{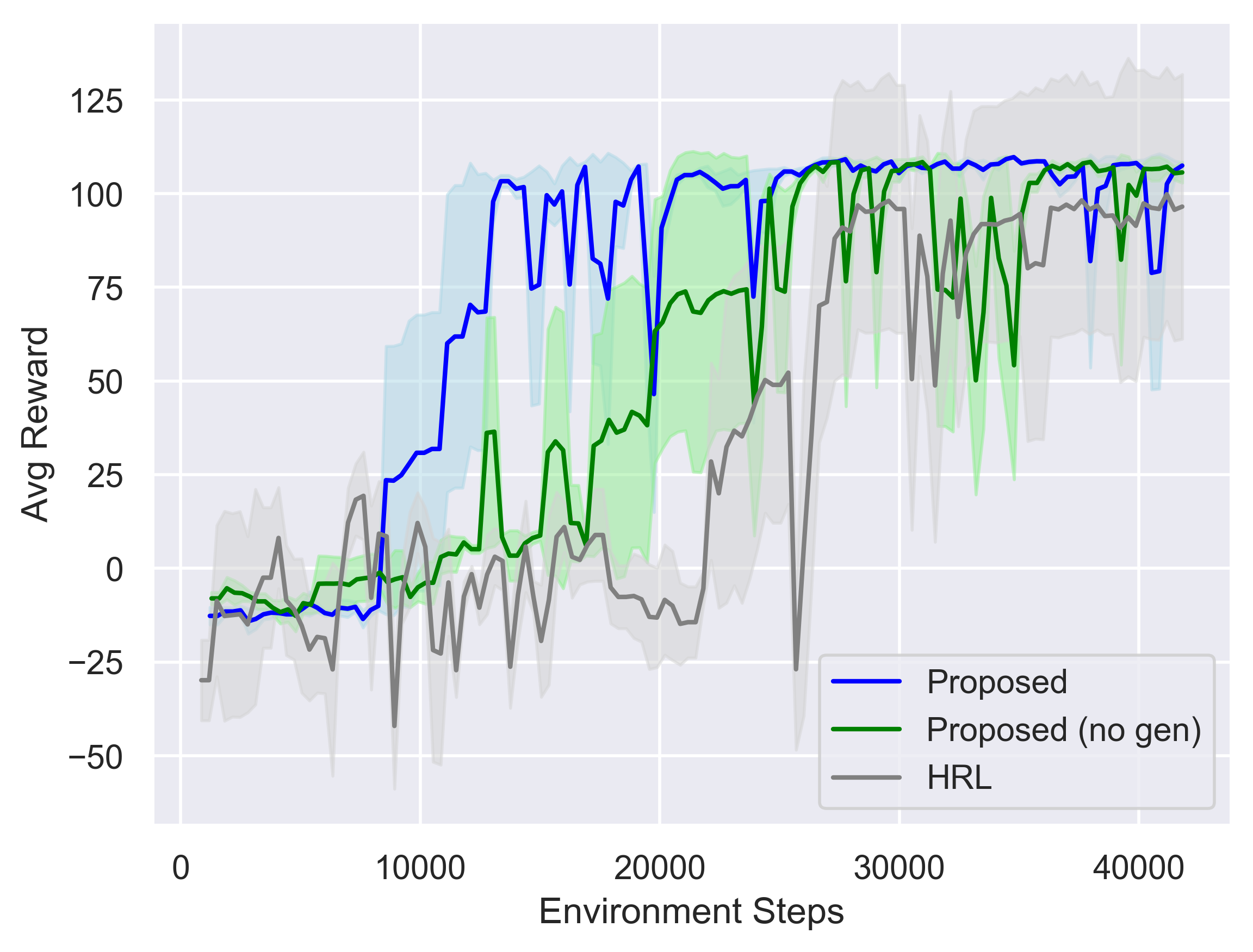}
    \label{fig:room2_5}
}
\subfigure[Test 2 Curve]{
    \includegraphics[width=1.39in]{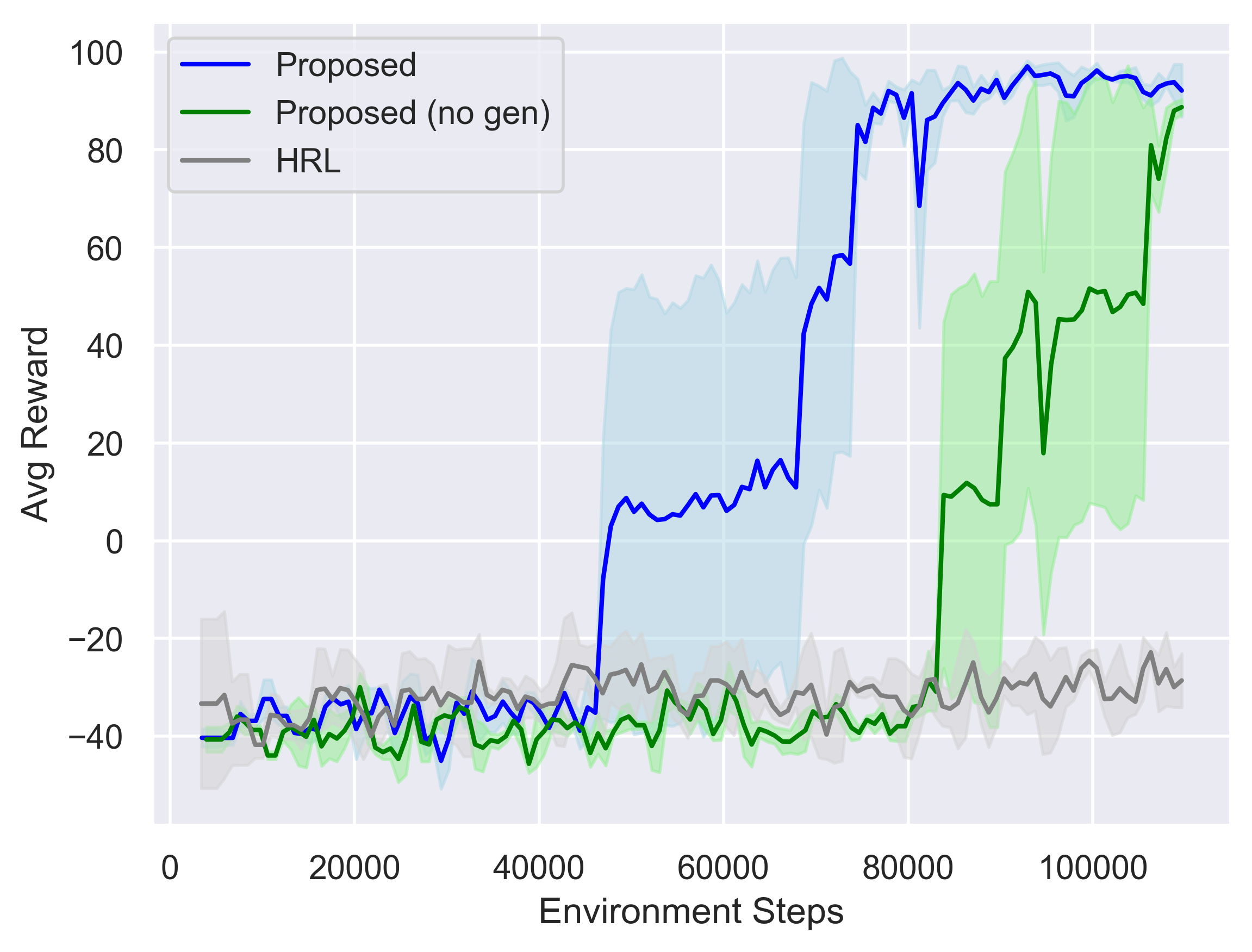}
    \label{fig:room2_6}
}
\vspace*{-5pt}
\caption{Testing Environments and Compositional Generalization. Three testing environments are shown in the first row, and performance comparisons are in the second row. The hierarchical DQN did not solve test 3 in 200 episodes.}
\label{fig:room2}
\vspace*{-5pt}
\end{figure*}

\vspace*{-5pt}
The room environment is a classical testbed for hierarchical RL, used in many previous HRL papers \citep{gopalan2017planning,le2018hierarchical,abel2020value}. It is to navigate a robot to the target room. In this work, in order to complicate the symbolic state, we add locks and keys in various colors and place them in different rooms. The training map is shown in Figure \ref{fig:room1_1}, and testing maps are in Figure \ref{fig:room2_2} and \ref{fig:room2_3} with more rooms and pairs of locks and keys added. In addition, the robot has no prior knowledge on state transitions or properties of the environment. More details of the experiments on room environment are presented in Section \ref{sec:app_room} in Appendix.

{\bf Setup} The training map consists of $17\times17$ grids, evenly partitioned into $4\times4$ rooms, shown in Figure \ref{fig:room1_1}. Every room occupies $3\times3$ grids, and adjacent rooms are separated by wall segments (yellow blocks). Some pairs of adjacent rooms are connected by corridors. Some rooms have keys, and some corridors are blocked by locks. The lock can only be opened by the key in the same color. And the robot has to open several locks before reaching the target room. 
In addition, the robot can only observe the current room and has no prior knowledge on the connectivity of other rooms or locations of locks and keys. Hence this environment defines a POMDP problem \citep{sutton1999between,kaelbling1998planning}. 

{\bf Experiment Result} The training experiment is to compare the sample efficiency of the proposed method with baselines including symbolic deep RL (SDRL) \citep{lyu2019sdrl} and hierarchical DQN \citep{kulkarni2016hierarchical}, on the map in Figure \ref{fig:room1_1}. The testing experiments are designed to verify the generalization capability of the proposed method, where the baselines are hierarchical DQN and the proposed method without using the symbolic transition model learned during training. 


The symbolic state transition model is trained by the transition experience in the symbolic MDP $\hat{\mathcal{M}}$. The symbolic transitions are partitioned into preconditions and effects of achieving subgoals. By leveraging $\partial$ILP introduced in Section \ref{sec:ilp}, with predicates specified in Table \ref{tab:room_preidcate} in Appendix, we can train and learn logic rules describing the preconditions as below,
\begin{itemize}[leftmargin=*]
    \vspace*{-5pt}
    \item 1: ReachRoom(Y)$\longleftarrow$CurAct(X,Y), Connect(X,Y)
    \vspace*{-5pt}
    \item 2: ReachRoom(Y)$\longleftarrow$CurAct(X,Y), Lock(X,Y,C), hasKeyColor(C)
    \vspace*{-5pt}
\end{itemize}
where rule 1 denotes going from room X to Y through a corridor, and rule 2 denotes the case of opening a lock with the right key. The learned logic rules for effects are shown in the following,
\begin{itemize}[leftmargin=*]
    \vspace*{-5pt}
    \item 1: visited(X)$\longleftarrow$ReachRoom(X)
    \vspace*{-5pt}
    \item 2: hasKeyColor(C)$\longleftarrow$ReachRoom(X), RoomHasKeyColor(X,C)
    \vspace*{-5pt}
\end{itemize}
where rule 1 presents the general effect of reaching a room, and rule 2 shows the effect of reaching a room that contains a key.

The map used in the training experiment is shown in Figure \ref{fig:room1_1}, and the comparison of the proposed method with baselines on sample efficiency is shown in Figure \ref{fig:room1_2}. 
We can see that the proposed method is approximately 40\% more sample efficient than regular HRL, and 30\% more efficient than SDRL. 
Moreover, the subtask success rate is shown in Figure \ref{fig:room1_3} for two different number of trials for each subtask $K$ (in Line 9 of Algorithm \ref{alg}). It is observed that the subtask success rate can quickly approach 1, but it is not equal to 1 because the high-level agent can always select some infeasible subtasks with non-zero probability due to the $\epsilon$-greedy strategy.

We then evaluate the capability of generalization of the proposed method, where the training is on the map in Figure \ref{fig:room1_1} and the testing is on two different maps in Figure \ref{fig:room2}. The corresponding results are shown in Figure \ref{fig:room2_5} and \ref{fig:room2_6}. We can see that the symbolic transition model learned in the training can accelerate learning in testing experiments, verifying the generalization capability of the proposed method.

\vspace*{-5pt}
\subsection{Robot Navigation}
\vspace*{-5pt}
In this environment, the robot is required to visit a number of circles with different colors, with a specified order. We modify the classical OpenAI's Safety Gym \cite{ray2019benchmarking}. The high-level agent is to select a sequence of circles to visit. The visiting of circles has to satisfy some constraints on ordering specified by the environment, initially unknown to the agent. In the low level, the robot tries to reach circles in the sequence one by one. We use tabular Q learning \citep{watkins1992q} and PPO \citep{schulman2017proximal} for high and low level learning respectively. 
There are 4 circles in 4 colors in training environment, and there are 8 (6) circles in 4 (6) colors in first (second) testing environment. More details of this environment and experiments are presented in Section \ref{sec:app_robot} in Appendix.

\begin{figure}
\centering
\subfigure[Game Screen]{
    \includegraphics[width=1.25in]{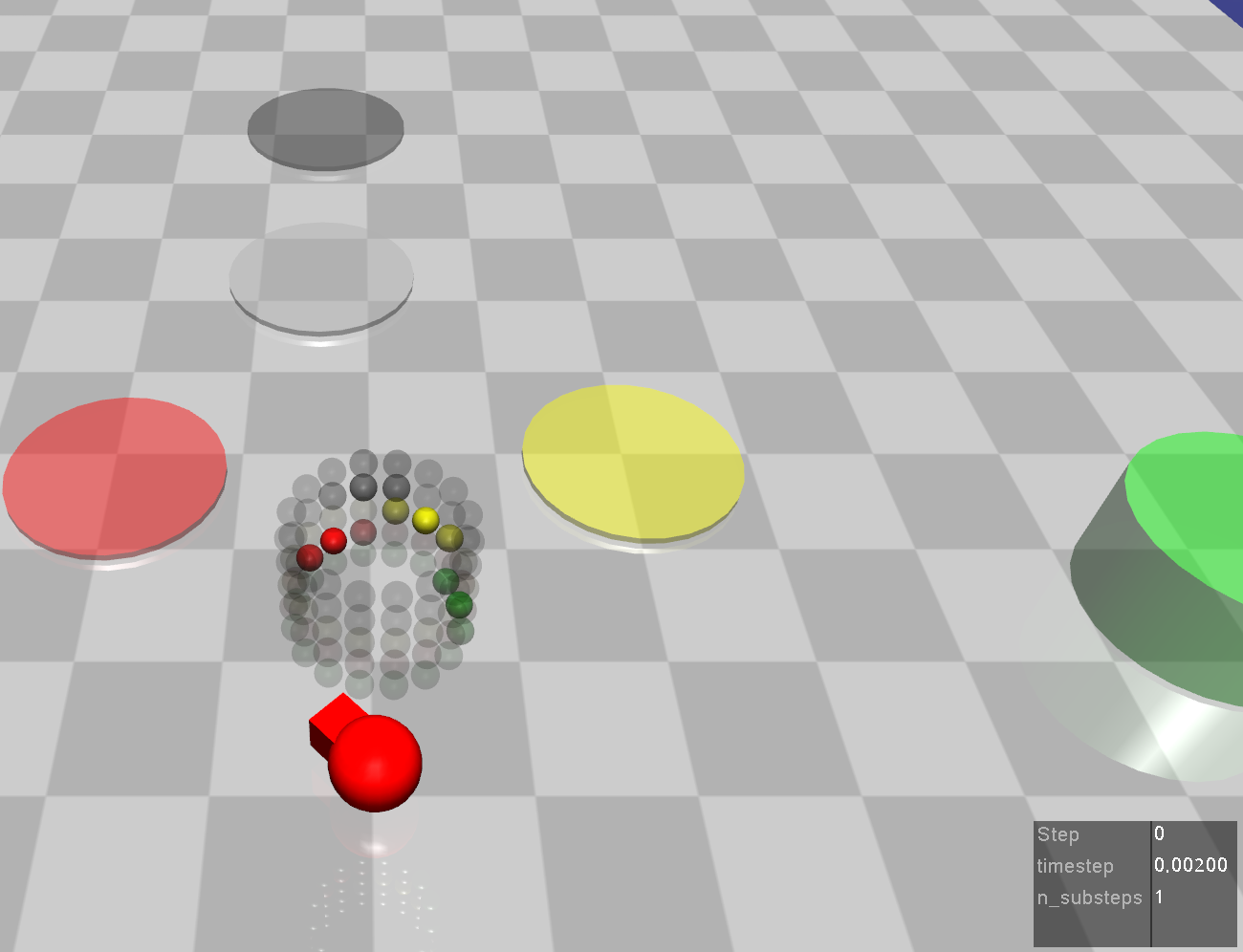}
    \label{fig:robot_screen}
}
\subfigure[Training]{
    \includegraphics[width=1.25in]{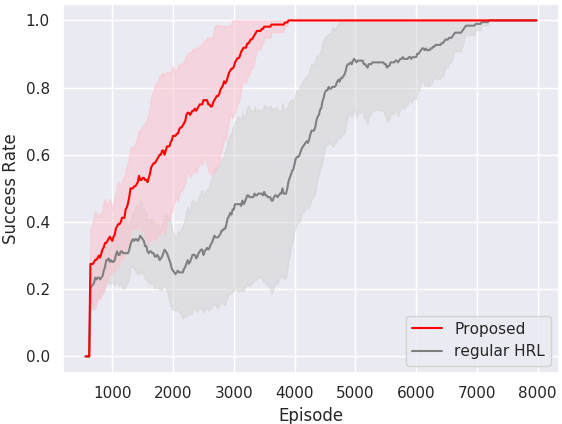}
    \label{fig:robot_training}
}
\subfigure[Test 1]{
    \includegraphics[width=1.25in]{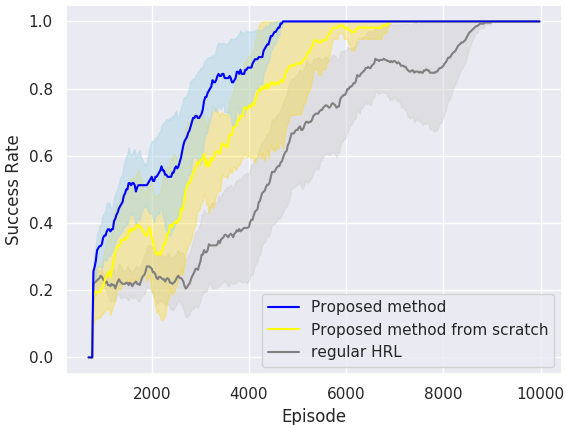}
    \label{fig:robot_test1}
}
\subfigure[Test 2]{
    \includegraphics[width=1.25in]{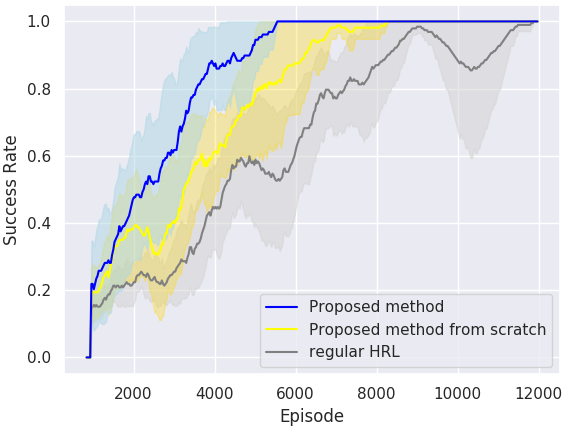}
    \label{fig:robot_test2}
}
\vspace*{-5pt}
\caption{Robot navigation environment. (a) Example game screen. (b) Performance on training environment. (c) Performance on the first testing environment. (d) Performance on the second testing environment. The success rate is the proportion of episodes which successfully finish the game in past 32 testing episodes.}
\label{fig:robot}
\vspace*{-10pt}
\end{figure}

{\bf Setup} In this environment, there is a 2D plane surrounded by walls which contains a number of circles in different colors corresponding to subgoals. The circles and the robot are randomly placed on the plane at the start of each episode and the robot has to visit circles in different colors in a specified order. An example screen is shown in Figure \ref{fig:robot_screen}. We use Safety Gym’s Point robot with actions for steering and forward/backward acceleration. The robot observes lidar information towards the circles and other sensory data (e.g., accelerometer, velocimeter). The target of the robot is to traverse circles in all colors with minimum steps, i.e., at least one circle in every color. 

{\bf Experiment Results} We use ILP to learn logic rules describing preconditions and effects of achieving subgoals. With predicates defined in Table \ref{tab:robot_predicate} in Appendix, the learned preconditions of subgoals are listed as below.
\begin{itemize}[leftmargin=*]
    \vspace*{-5pt}
    \item 1: AchieveObj(Y)$\longleftarrow$CurAct(X,Y), Connect(X,Y), isRed(Y)
    \vspace*{-5pt}
    \item 2: AchieveObj(Y)$\longleftarrow$CurAct(X,Y), Connect(X,Y), isYellow(Y), visitedRed()
    \vspace*{-5pt}
    \item 3: AchieveObj(Y)$\longleftarrow$CurAct(X,Y), Connect(X,Y), isGrey(Y)
    \vspace*{-5pt}
    \item 4: AchieveObj(Y)$\longleftarrow$CurAct(X,Y), Connect(X,Y), isBlack(Y), visitedGrey()
    \vspace*{-5pt}
\end{itemize}
The rules 2 and 4 above reflect the constraints on visiting yellow and black cirles. The effects of achieving subgoals are shown as below.
\begin{itemize}[leftmargin=*]
    \vspace*{-5pt}
    \item 1: visitedRed()$\longleftarrow$AchieveObj(X), isRed(X)
    \vspace*{-5pt}
    \item 2: visitedYellow()$\longleftarrow$AchieveObj(X), isYellow(X)
    \vspace*{-5pt}
    \item 3: visitedGrey()$\longleftarrow$AchieveObj(X), isGrey(X)
    \vspace*{-5pt}
    \item 4: visitedBlack()$\longleftarrow$AchieveObj(X), isBlack(X)
    \vspace*{-5pt}
\end{itemize}

As shown in Figure \ref{fig:robot_training}, the proposed method is around 40\% faster than the hierarchical baseline, showing the improvement due to the learned transition model. That is because, even though policies of options may not be well learned (subtask success rate $<0.9$) in early episodes, the optimal sequence of subgoals can be discovered based on symbolic transition model, as long as the connectivity of circles and symbolic transitions is discovered. In Figure \ref{fig:robot_test1} and \ref{fig:robot_test2}, comparing the blue and yellow curves, the method utilizing the symbolic transition model learned during training can improve the sample efficiency significantly, showing the generalization capability of the proposed method to environments with more circles and colors.

\vspace*{-5pt}
\section{Conclusion}
\vspace*{-5pt}
In this work, we proposed a new hierarchical framework for symbolic RL. In order to improve data efficiency and interpretability, leveraging the power of inductive logic programming (ILP), we learn a symbolic transition model in symbolic states. Therefore, the high-level agent can also conduct learning over this learned model in addition to the real environment, which saves a lot of samples and improves data-efficiency. The transition rules induced by ILP can also reveal the high-level working mechanism of the environment, introducing task-level interpretability. In $\partial$ILP method, in order to break the limit on the number of predicates in the body of clauses, we also integrate refinement into the original $\partial$ILP, so that many meaningless clauses can be avoided and rule templates are not required.


\bibliography{reference}

\renewcommand\thesection{\Alph{section}}

\onecolumn
\newpage

\section{Appendix}
\subsection{Refinement Operation}
\label{sec:app_refinement}
In the proposed method, we integrate $\partial$ILP with refinement operation \citep{cropper2021learning} which is used to generate clauses in $\mathcal{C}$. The learning of $\partial$ILP model is to train weights to select optimal clauses from $\mathcal{C}$, so that real transition experiences can be explained by selected logic rules. Here we consider two types of refinement operators to generate clauses, the addition of predicates into the body of clauses and replacement of variables by other variables. For example, given the general clause ReachRoom(Y)$\longleftarrow$CurAct(X,Y), the refined ones could be ReachRoom(Y)$\longleftarrow$CurAct(X,Y), Connect(X,Y), and ReachRoom(Y)$\longleftarrow$CurAct(Y,Y), generated by two types of operators, adding a predicate and replacing a variable. 

Omitting rigorous definitions in logic theory, in this method we generate clauses in $\mathcal{C}$ by refining general clauses incrementally. We only update clauses in $\mathcal{C}$ when any symbolic transition unpredictable by the transition model is encountered. According to Section \ref{sec:framework}, based on the definition of ILP \citep{koller2007introduction}, successful transitions correspond to positive examples while unsuccessful transitions are regarded as negative examples. Hence, given any symbolic state transition $(\hat{s},p,\hat{s}')$, the transition model is trained to predict the achievement of subgoal $p$ in successful transitions and not to deduce the subgoal $p$ in unsuccessful transitions. If a successful transition cannot be explained by the transition model where subgoal $p$ is predicted not achieved, it shows that current clauses in $\mathcal{C}$ are not enough, and we have to add new clauses into $\mathcal{C}$ by refining the most general clause ReachRoom(Y)$\longleftarrow$CurAct(X,Y), until this unexplained transition can be predicted by newly added clauses. However, if any unsuccessful transition $(\hat{s},p,\hat{s}')$ is not precluded where the subgoal $p$ is predicted to be achieved by some clause $c\in\mathcal{C}$, then we refine the clause $c$ until the subgoal $p$ cannot be deduced by clause $c$ given $\hat{s}$, and remove the original clause $c$ from $\mathcal{C}$.

\subsection{Algorithm}
\label{sec:alg}
The details of the proposed framework is shown in Algorithm \ref{alg}.
\begin{algorithm}
\caption{The Proposed Model-based Hierarchical Reinforcement Learning}
\label{alg}
\begin{algorithmic}[1]
\REQUIRE 
Predicates $\mathcal{P}$, labeling function $L$, termination function $\beta$, replay buffer $\mathcal{B}$, exploration parameter $\epsilon>0$, maximum steps in one episode $T_{\text{max}}$, symbolic transition model $\hat{T}_{H, \bm{\phi}}$, the mapping $L_G(s)$ from primitive state to currently achieved subgoal, the set of clauses $\mathcal{C}$ from which the symbolic transition rules are learned; \\
\STATE Initialize the high-level Q network $Q_h$ randomly;
\FOR{$e=1,\ldots,$}
\STATE $l\leftarrow 0$
\STATE Reset the environment and observe the initial state s;
\STATE Obtain the current symbolic state $\tilde{s}\leftarrow L(s)$
\WHILE{the goal has not been reached or $l<T_{\text{max}}$}
\STATE Choose subgoal $p\in\mathcal{P}_G$ by $\epsilon$-greedy according to $Q_h$
\IF{e is even}
\STATE Starting at $s$ as initial state, the subtask solver tries to achieve $p$ for $K$ times.
\STATE Update success ratio $t$ and number of trials $n$ for the subtask $(L_G(s), p)$
\STATE Compute the extrinsic reward $r_e(L_G(s), p)$ as \eqref{extrinsic}
\STATE $s'\leftarrow s$ and $\hat{s}'\leftarrow L(s)$
\IF{any successful trials}
\STATE Assign $s'$ by the last state of certain successful trial.
\STATE Update $\hat{s}'\leftarrow L(s')$ and $s\leftarrow s'$
\ENDIF
\STATE Store the transition tuple $(\hat{s}, \hat{a}, \hat{s'}, r_e)$ into replay buffer $\mathcal{B}$
\STATE If $(\hat{s},p,\hat{s}')$ cannot be explained by model $\hat{T}_H$, we use refinement operation to update $\mathcal{C}$
\ELSE
\STATE Predict the reward $r_e$ and next state $\hat{s}'$ by transition model $\hat{T}_{H,\bm{\phi}}$
\ENDIF
\STATE Update Q network $Q_h(\cdot,\cdot)$ 
\STATE $l\leftarrow l+1$
\STATE $\hat{s}\leftarrow \hat{s}'$
\ENDWHILE
\STATE Randomly sample a minibatch of transitions $\{(\hat{s}, \hat{a},\hat{s}',r_e)\}$ from $\mathcal{B}$
\STATE Update the symbolic transition model $\hat{T}_{H,\bm{\phi}}$ with the minibatch and clauses in $\mathcal{C}$, by using the ILP method in Section \ref{sec:ilp}
\ENDFOR
\end{algorithmic}
\end{algorithm}

\subsection{Experiments in Room Environment}
\label{sec:app_room}
In this section, more details of the experiments on room environment are presented.

{\bf Predicates} In this environment, we define the subgoal predicates as $\mathcal{P}_G:=$\{ReachRoom(X), where X$=1,\ldots,N$\} where $N$ is the number of rooms on the map. The environmental properties are described by predicates in $\mathcal{P}_C:=$\{Connect(X,Y), RoomHasKeyColor(X,C), Lock(X,Y,C), where X,Y$=1,\ldots,N$, C$=1,\ldots,M$\} where $M$ denotes the number of key colors on the map, variables $X,Y$ refer to indices of rooms, C refer to the index of the color. The environmental events are contained in $\mathcal{P}_E:=$\{visited(X), hasKeyColor(C), where X$=1,\ldots,N$, C$=1,\ldots,M$\}. The specific definitions of these predicates are listed in Table \ref{tab:room_preidcate} in Appendix. Another predicate CurAct(X,Y) denotes the fact that the robot is currently in room X and intends to go to room Y.

In the environmental MDP used in the low-level portion, every movement of the agent incurs a reward of $-1$, encouraging the agent to follow the shortest path. The reward of reaching the target is $100$. There are no rewards for other situations, making the environmental rewards sparse. The robot can only hold one key at a time. 

{\bf Experiment Details} In the training experiment, the first baseline is regular HRL \citep{kulkarni2016hierarchical} where Q learning is adopted in both high and low levels. The second baseline is SDRL \citep{lyu2019sdrl} whose the action description is reformulated, where everything about state transitions are removed. We adopt $\epsilon$-greedy for action selection in the high level, where $\epsilon$ is linearly decreasing from $0.3$ to $0.03$. The performance comparison is shown in Figure \ref{fig:room1_2}. The high-level agent can also interact with the learned symbolic transition model, reducing the sample complexity significantly. And the logic rules in the symbolic transition model can make the users understand the high-level operating mechanism of the environment, improving the interpretability. In the low level, we use deep Q learning for all the baselines and the proposed method, where the Q network is realized by a 2-layer MLP with 32 neurons and ReLU activation function in every layer. 

In testing experiments, the first map in Figure \ref{fig:room2_2} has more locks to open than the training map, and the second one shown in Figure \ref{fig:room2_3} has larger size with a deceptive lock leading to a dead-end. The baselines are regular HRL and the proposed method which learns the transition model from scratch in the testing. Here the Q function learned in training cannot be used directly, since the room connectivity has changed in the testing maps. However, we can see that the symbolic transition rules still hold in the testing maps, even though there are colors and rooms. Hence the high-level agent can still utilize the symbolic transition model learned during training to reduce sample complexity. For example, when the locations of green key and lock are identified, the robot with learned transition model can quickly figure out to first pick up the key and then open the lock. Thus, as shown in Figure \ref{fig:room2}, even though the map is changed in testings, the proposed method can still solve testing maps faster than baselines including the proposed method without using the transition model learned during training, demonstrating the effect of the generalization capability of the symbolic transition model in the proposed method. Particularly, the regular HRL cannot solve Test 2 in Figure \ref{fig:room2_3} within 200 episodes while the proposed method can solve that within 100 episodes. 

\begin{table}
\footnotesize
\caption{Definitions of Predicates in Room Environment}
\vspace*{+5pt}
\centering
\footnotesize
\begin{tabular}{|c|c|}
\hline\hline
Name & Definition \\\hline
$\text{CurAct}(X, Y)$ & \tabincell{c}{The room $X$ the robot currently stays, and the intended subgoal $Y$} \\\hline
$\text{ReachRoom}(X)$ & The robot will come to room $X$ at next time step \\\hline
$\text{Connect}(X, Y)$ & Room $X$ and room $Y$ are connected by a corridor without any lock \\\hline
$\text{Lock}(X, Y, C)$ & There is a lock between room $X$ and $Y$ in color $C$ \\\hline
$\text{RoomHasKeyColor}(X, C)$ & Room $X$ has a key in color $C$ \\\hline
$\text{Visited}(X)$ & Room $X$ has been visited by the robot in current episode \\\hline
$\text{hasKeyColor}(C)$ & The robot has obtained the key in color $C$ \\\hline
\end{tabular}
\label{tab:room_preidcate}
\end{table}

\subsection{Robot Navigation}
\label{sec:app_robot}
In this section, we introduce more details about the experiments on the robot navigation. In the high level, the constraints on visiting circles are specified by the environment, where the precondition of visiting yellow (black) circle is that the red (grey) circle has been visited. These constraints are unknown to the agent initially. In addition to the extrinsic reward, the reward of finishing the game successfully is $10$.

In the low level, for every circle, a specific option is trained to reach that circle as the subgoal. The policy network of options consists of a two-layer MLP with 64 neurons in each layer. Based on intrinsic rewards in \eqref{intrinsic} with $\eta=1$, we use PPO algorithm \citep{schulman2017proximal} to learn the option policy. For any circle X and Y, the predicate Connect(X,Y) is set to be true, as long as the number of steps of going from circle X to Y is less than $T_s=300$. More hyperparameters are listed in Table \ref{tab:hyp_robot}.

{\bf Predicates} Here the subgoal predicates are to achieve circles indexed from $1$ to $N$, i.e., $\mathcal{P}_G:=$\{AchieveObj($X$), $X=1,\ldots,N$\}. The environmental properties include the color of all the circles and connectivity among them, i.e., $\mathcal{P}_C:=$\{isRed(X), isYellow(X), isGrey(X), isBlack(X), Connect(X,Y), $\forall$ X,Y=$1,\ldots,N$\}. The environmental event predicates in $\mathcal{P}_E$ include \{visitedRed(), visitedYellow(), visitedGrey(), visitedBlack()\}, denoting whether a circle in every color has been visited or not. We also use predicate CurAct(X,Y) to denote the robot's current circle X and intended circle Y to go. The definitions of predicates are listed in Table \ref{tab:robot_predicate}. 

{\bf Experiment Details} In the training environment, there are 4 circles in the plane which are in red, yellow, grey and black, with some visiting constraints. In testing environments, there are more circles in more colors designed to verify the generalization capability of the proposed method, but the constraints in the training experiments still hold. The first testing environment has 8 circles in 4 colors, and the second one has 6 circles in 6 colors, where 4 colors in training (red, yellow, grey and black) still appear in both testing environments. The baseline adopted here is the regular hierarchical RL method where PPO \citep{schulman2017proximal} is applied in the low level and tabular Q learning \citep{watkins1992q} is used in the high level. In the testing environment, another benchmark is the proposed method in which the symbolic transition model is trained from the scratch, without using the model learned in the training.

\begin{table}
\footnotesize
\caption{Definitions of Predicates in Robot Navigation}
\vspace*{+5pt}
\centering
\footnotesize
\begin{tabular}{|c|c|}
\hline\hline
Name & Definition \\\hline
CurAct(X,Y) & \tabincell{c}{The robot is currently in circle X and intends to go to circle Y} \\\hline
AchieveObj(X) & The robot will come in circle X successfully \\\hline
Connect(X,Y) & The robot can go from circle X to Y with steps less than step limit $T_s$ \\\hline
isRed(X), isYellow(X), $\ldots$ & The color of circle X \\\hline
visitedRed(), visitedYellow(), $\ldots$ & Any circle in certain color has been visited \\\hline
\end{tabular}
\label{tab:robot_predicate}
\end{table}

\begin{table}
\footnotesize
\caption{Hyperparameters in Robot Navigation}
\vspace*{+5pt}
\centering
\footnotesize
\begin{tabular}{c|c}
Hyperparameter & Value \\\hline\hline
Number of subtask trials ($K$) & 10 \\
Timeout step $T_{\text{max}}$ & 1000 \\
Minibatch size & 16 \\
Discount factor ($\gamma$) & 0.99 \\
Learning rate & $3\times10^{-4}$ \\
GAE-$\lambda$ in PPO & $0.95$ \\
Entropy coefficient & $0.003$ \\
Value loss coefficient & $0.5$ \\
Gradient clipping & $1.0$ \\
PPO clipping ($\epsilon$) & $0.2$
\end{tabular}
\label{tab:hyp_robot}
\end{table}

\end{document}